\documentclass[10pt,twocolumn,letterpaper]{article}

\usepackage{cvpr}
\usepackage{times}
\usepackage{epsfig}
\usepackage{graphicx}
\usepackage{amsmath}
\usepackage{amssymb}
\usepackage{cvpr}
\usepackage{times}
\usepackage{epsfig}
\usepackage{graphicx}
\usepackage{amsmath}
\usepackage{amssymb}
\usepackage{mathtools}
\usepackage[greek, english]{babel}
\usepackage{multirow}
\usepackage{enumitem}
\usepackage{caption}
\usepackage{subcaption}
\usepackage{cvpr}
\usepackage{times}
\usepackage{epsfig}
\usepackage{graphicx}
\usepackage{amsmath}
\usepackage{amssymb}
\usepackage{mathtools}
\usepackage{multirow}
\usepackage{enumitem}
\usepackage{caption}
\usepackage{subcaption}
\usepackage{sidecap}

\usepackage[capbesideposition=outside,capbesidesep=quad]{floatrow}
\makeatletter
\renewcommand{\paragraph}{%
  \@startsection{paragraph}{4}%
  {\z@}{1ex \@plus 1ex \@minus .2ex}{-1em}%
  {\normalfont\normalsize\bfseries}%
}
\makeatother
\usepackage[pagebackref=true,breaklinks=true,letterpaper=true,colorlinks,bookmarks=false]{hyperref}

\cvprfinalcopy 

\def\cvprPaperID{2459} 

\ifcvprfinal\fi
\begin{document}

\title{Texture Synthesis Guided Deep Hashing for Texture Image Retrieval}

\author{Ayan Kumar Bhunia\textsuperscript{1} \hspace{.1cm} Perla Sai Raj Kishore\textsuperscript{2} \hspace{.1cm} Pranay Mukherjee\textsuperscript{2} \hspace{.1cm} Abhirup Das\textsuperscript{2} \hspace{.1cm} Partha Pratim Roy\textsuperscript{3}\\
\textsuperscript{1}Nanyang Technological University, Singapore
\hspace{.1cm}  \\ \textsuperscript{2} Institute of Engineering \& Management, India  \hspace{.1cm} \textsuperscript{3} Indian Institute of Technology Roorkee, India\\
{\tt\small \textsuperscript{1}ayanbhunia@ntu.edu.sg }
}

\maketitle
\ifcvprfinal\thispagestyle{empty}\fi

\begin{abstract}
   With the large scale explosion of images and videos over the internet, efficient hashing methods have been developed to facilitate memory and time efficient retrieval of similar images. However, none of the existing works use hashing to address texture image retrieval mostly because of the lack of sufficiently large texture image databases. Our work addresses this problem by developing a novel deep learning architecture that generates binary hash codes for input texture images. For this, we first pre-train a Texture Synthesis Network (TSN) which takes a texture patch as input and outputs an enlarged view of the texture by injecting newer texture content. Thus it signifies that the TSN encodes the learnt texture specific information in its intermediate layers. In the next stage, a second network gathers the multi-scale feature representations from the TSN's intermediate layers using channel-wise attention, combines them in a progressive manner to a dense continuous representation which is finally converted into a binary hash code with the help of individual and pairwise label information. The new enlarged texture patches from the TSN also help in data augmentation to alleviate the problem of insufficient texture data and are used to train the second stage of the network. Experiments on three public texture image retrieval datasets indicate the superiority of our texture synthesis guided hashing approach over existing state-of-the-art methods.
\end{abstract}

\section{Introduction}

Recent times have seen a huge explosion of digital images over the internet which has made large image databases prevalent.
Given any database one might want to search for images semantically using a query image. 
Content Based Image retrieval (CBIR) explored in \cite{4711909, 6777560, liu2011image, Magliani:2018:ART:3243394.3243686, 6175124, ojala1996comparative}
provides a solution to the above mentioned problem by retrieving a set of similar images by the measure of a similarity metric between the feature representations of the query image and the member images of the database. Thus a general image retrieval pipeline consists of two steps : first, characterization of each image by rich discriminative features and second, performing a similarity search by some metric using these features to retrieve similar images.

Feature representations for images are usually continuous-valued and thus running a nearest neighbour search on these representations for retrieval turns out to be very slow and computationally inefficient, especially for real-time  applications in mobile devices or in databases with millions of images.
Hence, there is a need to optimize this naive search technique under the constraints of both space and time. 
Hashing \cite{wang2014hashing} is one such state-of-the-art technique for Approximate Nearest Neighbor (ANN) search \cite{6909695, magliani2018efficient, muja2014scalable} used due to faster retrieval speeds and reduced memory footprint. 
It involves learning a hash function for encoding continuous-valued image descriptors to compact binary codes while preserving their similarity and discriminative properties.  
The similarity between two such hash codes can be easily computed by their Hamming distance with the simple XOR operation. 
Recently, performance of traditional hashing methods have been bettered by deep hashing techniques like 
\cite{ 10.1007/978-3-030-01228-1_21, cao2017hashnet, he2017hashing, lai2015simultaneous, lai2016instance, lin2015deep, zhang2015bit, zhao2015deep} which employ deep neural networks for learning the hash function.  
Consequently, our work uses deep hashing to address a special kind of image retrieval task called Texture Image Retrieval.
 
Texture image retrieval may be defined as a type of CBIR which aims at searching for images having texture-patterns semantically similar to that of the query image.  
Texture being a low level visual attribute of an image's surface acts as a representative of the surface's roughness and also provides useful visual cues about the object's identity.
Texture image retrieval has a variety of applications such as in digital library, multimedia web search, multimedia storage system and query-based video investigation. 
However, this task is very challenging for a couple of reasons. 
Firstly, due to the lack of large scale texture databases, this task is much less explored in the context of deep learning as compared to ordinary image retrieval. 
Secondly, for image retrieval tasks in general, the closeness among images is usually governed by the similarity among their high level features. 
However, in texture image retrieval where similarity is governed by texture patterns, similar semantics become difficult to capture owing to the fact that texture is a low level visual attribute.

In our work, we introduce a deep hashing framework for texture image retrieval guided by a Texture Synthesis Network (TSN) \cite{xian2017texturegan}. 
We extract information at various levels from the intermediate layers of a pre-trained TSN and combine these multi-scale activations using channel-wise attention in a progressive manner to generate a powerful set of feature descriptors for texture images.
The example-based texture synthesis approach aims at generating a texture image, double the size of input that faithfully captures all the visual properties of the input by preserving the large-scale structural features, the natural appearance and the spatial variation of local patterns. 
For this we use a generative adversarial network (GAN) where the generator aims to synthesize a larger image with an expanded view of the given input texture and the discriminator aims at comparing and classifying the generated textures with the corresponding ground truth images.
Although this might resemble image super-resolution \cite{ledig2017photo}, they are different in various ways. 
Image super-resolution simply aims at enhancing the quality of the given image by increasing its resolution and sharpening it, whereas, texture synthesis aims at expanding the view of the given texture patch by injecting additional content consistent with the given input. 
Since the generator is able to synthesize larger textures from smaller patches, it is evident that all of the texture related information is recorded in the intermediate layers of the generator network. This information is used as the key source in our work for generating good descriptors for texture images. 

Channel-wise attention helps us to combine activation maps from various layers of the pre-trained TSN in a selective manner and reduce noise, such that channels with more information are given greater weightage than noisy and less informative ones. 
These continuous valued features are finally hashed to generate compact binary coded representations of the images.
In addition to images from existent texture datasets, we also make use of images generated from our TSN to alleviate the problem of insufficient texture data for training which significantly improves the performance of our deep neural network.

In this paper, we make the following novel contributions: 
\begin{itemize}[wide, labelwidth=!, labelindent=0pt, nosep, topsep=0pt]
\item We introduce a deep hashing network for texture image retrieval, which uses the rich texture information recorded in the intermediate layers of a pre-trained TSN, filters them using channel-wise attention and then combines these multi-scale feature representations in a progressive manner to get a powerful and robust set of feature descriptors for texture images. We finally learn a hash function to map these continuous valued feature descriptors to dense binary codes. 
\item To the best of our knowledge, our work is the first to introduce an end-to-end deep neural network and hashing for texture image retrieval. Experimental results on various benchmark datasets exhibit a superior performance of our framework over existing methods. 
\end{itemize}

\begin{figure*}[t]
\begin{center}
\includegraphics[width=0.86\linewidth]{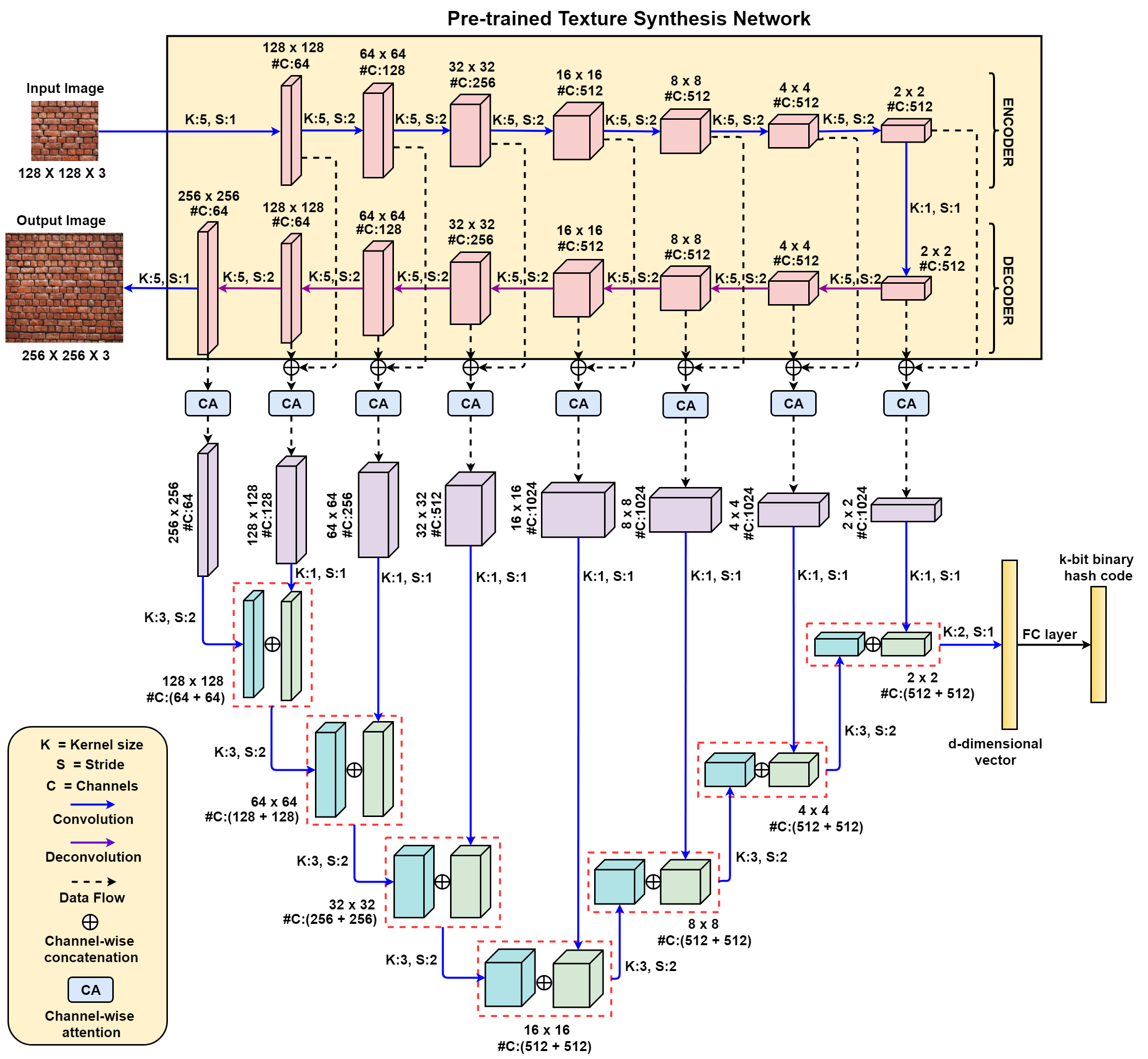}
\end{center}
   \caption{A $128 \times 128$ texture patch is passed as input to the pre-trained generator of the TSN which generates an expanded texture patch of size $256 \times 256$ as output. The corresponding activation maps of the same size from the encoder and decoder part of the generator are depth-wise concatenated and are passed through a channel-wise attention (CA) module to generate a filtered version of the intermediate features, followed by a $1 \times 1$ convolution operation. 
   We use separate CA modules for every pair of feature combination between encoder and decoder side of activation maps. The multi-scale features are then combined in a progressive manner using successive downsampling (by strided convolution) and depth-wise concatenation.}
\label{fig:short}
\end{figure*}

\section{Related Work}

Earlier methods for image retrieval relied on encoding techniques \cite{jegou2010aggregating, perronnin2007fisher, perronnin2010large} for aggregating local patches to build a global image representation.
Later methods such as \cite{gordo2017end, paulin2017convolutional} relied on convolutional neural networks (CNNs) for learning image features. 
Krizhevsky \etal \cite{krizhevsky2012imagenet} used feature information gathered by a classifier, pre-trained on large scale dataset like ImageNet, as the basis of developing suitable descriptors for instance-level retrieval tasks. 
A more recent work by Radenovic \etal \cite{radenovic2018fine} aims at generating good descriptors for image retrieval by fine-tuning CNNs on a large collection of unordered images in a fully automated manner. 
The authors of \cite{gordo2017end} proposed a large-scale dataset which they cleaned to produce less noisy training data for image retrieval. Furthermore, they improved the R-MAC feature descriptor \cite{tolias2015particular} and used a triplet loss for training it using a siamese network.

Few of the existing works compressed the descriptors to improve the storage requirements and retrieval efficiency at the cost of reduced accuracy. 
Both supervised \cite{gordoa2012leveraging} and unsupervised  \cite{jegou2012negative, perronnin2010large, radenovic2015multiple} compression techniques are used for this purpose.
The issue of compressing the feature descriptors for faster retrieval without compromising on accuracy is solved by hashing. 
Traditional hashing methods largely differ from the state-of-the-art deep learning based methods which learn the image representations and hash codes in an end-to-end manner.

The authors of \cite{xia2014supervised} proposed one of the first approaches to deep hashing called Convolutional Neural Network Hashing (CNNH) which used a two stage network for learning the image representations and the hash codes. 
However, their hash code learning process did not receive any kind of feedback from the learnt image representations.
Network in Network Hashing (NINH) \cite{lai2015simultaneous}  attempts to alleviate the problem faced in CNNH by learning the image representations and the hash codes at the same stage. 
Many other ranking-based deep hashing methods \cite{wang2016deep, yao2016deep, zhao2015deep} and pairwise label based deep hashing methods \cite{li2015feature, zhu2016deep} have been proposed in the recent years. 
The hashing part of our work draws inspiration from \cite{li2017deep} which proposes a single framework to learn the binary codes directly by using both pairwise and individual label information.

Although hashing has been proved to significantly increase the time and memory efficiency, it has still not been used in texture image retrieval. 
Earlier methods on texture image retrieval relied on handcrafted features like Discrete wavelet transform, Gabor features \cite{li2017color}, local binary pattern (LBP) \cite{liu2017fusion}, rotated wavelet filter, rotated complex wavelet filter and dual tree complex wavelet transform for feature extraction. 
Although LBP worked well for grayscale images, it performed poorly for colour images.
To alleviate this problem, the authors of \cite{liu2017fusion} introduced color-information feature (CIF) in addition to LBP to generate textural and color information for images under the settings of image retrieval and classification. 
Li \etal \cite{li2017color} use Gabor wavelets to decompose colour texture images into various dependencies, analyse them and then capture these dependencies with Gaussian copula models. 
A recent work by Banerjee \etal \cite{banerjee2018local} introduces a new feature descriptor called Local Neighbourhood Intensity Pattern (LNIP). We argue that in order to learn good feature descriptors for texture image retrieval, it is necessary to understand the underlying patterns and the low-level and high-level properties of the texture images. This can be done effectively by learning to generate a larger texture image from a smaller patch by injecting newer patterns in a consistent and non-repeating manner.

Texture synthesis aims exactly at this task and has been extensively researched for over the past two decades. Classical methods for texture synthesis include non-parametric approaches, such as \cite{efros1999texture, kwatra2003graphcut, lefebvre2006appearance, wei2000fast, wexler2007space}. Whereas recent approaches \cite{bergmann2017learning, jetchev2016texture, li2016precomputed, xian2017texturegan, zhou2018non} make use of deep learning based methods to significantly improve the quality of results.
One of the earliest methods for texture synthesis using deep neural networks include the work of Gatys \etal \cite{gatys2015texture} where they iteratively optimize an image to generate texture.  
Fang \etal \cite{fang2004textureshop} proposed a novel application of texture synthesis for editing images by applying texture to the surface of a photographed object.  
TextureGAN \cite{xian2017texturegan} uses a deep generative network that aims to synthesize textures in sketch images guided by user defined texture samples.
In \cite{zhou2018non}, Zhou \etal proposed a state-of-the-art generative approach for doubling the spatial extent of the given input patch by injecting newer texture content.

\section{Proposed Work}

This work introduces a novel deep hashing framework for texture image retrieval. 
We aim to generate good feature descriptors for texture images by transferring knowledge learned from the task of texture synthesis to texture image retrieval. 
Our framework can be divided into two stages: In the first stage, we train the TSN using an adversarial setup, where the generator learns to generate large texture images from smaller patches by injecting newer texture content. 
In the second stage, the pre-trained TSN is used for feature extraction of texture images for hashing. 
We combine the multi-scale activation maps from the intermediate layers of pre-trained generator network in a selective and efficient manner to generate a powerful set of feature descriptors. 
Finally, these continuous valued feature descriptors are hashed to compact binary vectors for efficient memory usage and faster retrieval. 
These binary valued vectors are then compared by their Hamming distance using a simple XOR operation. 
We also tackle the data insufficiency problem for training the deep network by using the newly generated texture patches from the TSN for learning the hash function.  
The details of our model and the training procedure is discussed in the subsequent sections.

\subsection{Texture Synthesis Network}

Generating large texture images from smaller patches forms the basis of generating powerful feature descriptors for the texture images in our work. 
A generative adversarial approach is employed for texture synthesis, with a convolutional encoder-decoder network as the generator and a fully convolutional classification network as the discriminator.

The generator  $G_{TSN}(.)$ takes in a smaller texture patch of size $K \times K$ as input and generates a larger texture image of size $2K \times 2K$ via adversarial expansion. 
In the encoder part of the generator, after every strided convolution operation, the spatial dimensions of the activation maps are halved and the number of channels are increased. 
Similarly in the decoder part, after each deconvolution operation, the spatial dimensions of the activation maps are doubled and the number of channels are reduced. 
The generator network is by purpose designed in a symmetric manner to allow easy combination of activation maps from the corresponding layers of the encoder and decoder network in the second stage of our framework. 
It is also to be noted that the generator is not perfectly symmetrical due to the extra deconvolution operation in the decoder part which is necessary for upsampling the image to size $2K \times 2K$. The details of the generator architecture has been shown in Figure \ref{fig:short}.  
The discriminator, $D_{TSN}(.)$, is a traditional convolutional classification network which takes in a large texture image (of size $2K \times 2K$) and classifies it as being generated by the generator (fake) or not (real). 
Our discriminator has an architecture similar to the one used in \cite{radford2015unsupervised}.

Along with the adversarial loss $L_{adv}$ \cite{goodfellow2014generative}, for simultaneous training of the generator and the discriminator, we follow \cite{zhou2018non} to use two additional loss terms which guide the generator to generate high quality expanded textures over time. 
We use $L_1$ loss between the image patch output by the generator and it's ground truth counterpart from the examples so as to provide a pixel-wise error supervision to the generator. 
Second, we use style loss $L_{style}$ \cite{gatys2016image} to enforce perceptual similarity between the generated image and the ground truth image. 
For this, we use Gram matrices \cite{gatys2016image} computed from the activations of convolutional layers of the pre-trained VGG-19 network. Therefore, the final loss equation is as follows:

\begin{equation}\label{l1_loss}
L_{total} =   L_{adv} + \gamma_{1} L_{style} + \gamma_{2} L_{L_{1}} 
\end{equation}

The pre-trained generator of this TSN is used for two purposes: first, to extract powerful feature descriptors for the texture images from its intermediate layers and second, to generate more data to train the second stage of the network for hashing.

\subsection{Channel-wise Attention}

In the second stage of our framework, we perform channel-wise concatenation of the corresponding activation maps of the same size from the encoder and decoder parts of the pre-trained TSN. 
The reason behind this concatenation is that we intend to combine both local and global information from the intermediate layers of the TSN. 
The activations from the encoder part hold the local texture-specific information extracted from the input patch, whereas the activations from the decoder part hold information about the inherent properties of the texture and a global class-specific representation of the same. 
Since the decoder part is mainly responsible for generating an expanded view of the input texture by injecting additional texture content, it can be interpreted that the information contained in the activation maps of the decoder part is more representative of the inherent global properties of the texture image. 
Every channel of any convolutional feature map has high response to some specific content of the image. Thereby it becomes important to assign more priority to those channels which contain more discriminative information in order to facilitate better feature learning. The Channel-wise Attention (CA) module in our framework is mainly used to encode this characteristic and in the process provides higher weightage to the channels carrying more texture specific information than the channels with lesser information. 
The output of the CA module is a tensor of same dimensions as that of input but channels weighted as per their scalar relevance value. 

An intermediate output of the CA module is a vector $A_{c}$  having dimension equal to the number of input channels  $C$, along which the attention operation is applied. 
Here, every $i^{th}$ element of $A_{c}$ represents the normalized contribution of the corresponding $i^{th}$ channel to the output of the attention module. 
$A_{c}$ is calculated as follows :

\begin{equation}\label{channelwise_attention}
 A_{c} =  softmax(W_{c} * q + b)
\end{equation}
In the above equation, $*$ operator represents the convolution operation, $W_{c}$ represents the convolutional filter, $b$ represents the bias factor and $q$ represents the vector of dimension $C$ obtained by performing global average pooling on every channel of the input to the channel-wise attention module. 
This normalized vector is finally multiplied with the original input tensor to get a filtered output. 
We use separate CA modules for every pair of feature combination between the encoder and the decoder side of activation maps. 
In our case, we use 8 different CA modules (see Figure \ref{fig:short}) in order to adaptively weigh different channels of the combined feature maps of sizes ranging from $256 \times 256$ till $2 \times 2$ by powers of $2$.

\subsection{Progressive Multi-scale Feature Combination}

Following the channel-wise attention operation, we intend to combine these multi-scale feature maps to get a final feature vector representation of the input texture patch. 
This progressive multi-scale feature combination module includes two operations: first, a $1 \times 1$ convolution operation upon the filtered output of the CA module and second, progressive combination through successive downsampling by strided convolution and depth-wise concatenation.
There are two main reasons behind the use of $1\times1$ convolution: feature mixing by combining channel-wise weighted feature maps and dimensionality reduction by reducing the number of channels equal to the depth of the previous layer for concatenation at a later stage. 
Let ${\mathrm{CA}}^{M}$ represent the output of a CA module applied to feature map of size $M \times M$, where M has values $256$, $128$, $64$, $32$, $16$, $8$, $4$, and $2$ . $1\times1$ convolution is applied to every $\mathrm{CA}^{M}$ , except where $M$ equals to $256$, resulting in a feature map $F_{1 \times 1}^{M}$. First, $\mathrm{CA}^{256}$ is passed through a strided convolutional layer giving $F_{strided}^{128}$ and then concatenated with $F_{1 \times 1}^{128}$ which is of same depth. 
In a similar fashion, the concatenated feature map is again passed through another strided convolutional layer by outputing $F_{strided}^{64}$ and similar operations are performed for successive downsampled feature maps to finally generate a $d$ dimensional vector. Therefore, it can be generalized from $M = 128$ onwards as follows:

\begin{equation}\label{FC1}
 F_{1 \times 1}^{M} = Conv_{1 \times 1}(\mathrm{CA}^{M}) 
\end{equation}

\begin{equation}\label{FC2}
F_{strided}^{M/2} = Conv_{strided} (F_{strided}^{M}  \oplus F_{1 \times 1}^{M}) 
\end{equation}
where, $Conv_{strided}(\cdot)$ is a strided convolution layer with kernel size $3 \times 3$  and stride $2$, and $Conv_{1 \times 1}(\cdot)$ is a $1 \times 1$ convolutional layer with number of filter equals to half of the depth of its input. A clearer and vivid visual description can be found in Figure \ref{fig:short}.

\subsection{Hash Function}

The previous sections involved generating a $d$ dimension feature descriptor for a given input texture patch. This section describes the use of hashing which converts these continuous feature representations to dense binary codes while preserving the semantic similarity between the images.

In the problem of hashing, given a set of $N$ images each of which have been described by the features of dimension $d$ the aim is to encode the feature matrix, $Z$, of dimension $d \times N$ into another matrix $H$ of dimension $k \times N$  such that every element of $H$ is either $1$ or $-1$. 
In other words every column of $Z$ that represents the image's complex features in $d$ dimensions is mapped to a binary feature representation of dimension $k$. 
Thus if we consider $ h(.) $ to be the hashing function, this process can be explained mathematically as $H_{i}  =  h(Z_{i})$, where $H_{i}$ represents the $i^{th}$ column of matrix $H$ and $Z_{i}$ represents the $i^{th}$ column of matrix $Z$.

Following \cite{li2017deep}, we make use of both pairwise label information and individual label information for supervising the hashing network. 
The similarity measure between two binary hash codes ($b_{i}$, $b_{j}$) is defined using the concept of Hamming distance $dist_{h}(\cdot,\cdot)$ and cosine similarity $\langle \cdot,\cdot \rangle$ as 
$dist_{h}(b_{i}, b_{j}) = \dfrac{1}{2}(k - \langle b_{i}, b_{j} \rangle)$. Since the inner product and Hamming distance are inversely related, we can use inner product to quantify similarity between the hash codes. 
Negative log likelihood reduces the Hamming distance between similar pair of images and increases the Hamming distance between dissimilar pairs. 
A simple linear classification network is used along side to exploit the label information directly and is based on the assumption that the learned binary codes should be good enough for classification as well. 
Therefore, our final loss function is a weighted summation of negative log likelihood function $J$ \cite{li2017deep} and classification loss $Q$ \cite{li2017deep}, and is given by :
\begin{equation}\label{J}
L_{hash} = J + \nu Q
\end{equation}
Since optimizing $L_{hash}$ is a discrete optimization problem, we follow the training and testing procedure proposed in \cite{li2017deep} for this purpose.

\section{Experiments}

\begin{figure*}[t]
\begin{center}
\includegraphics[width=1.05\linewidth]{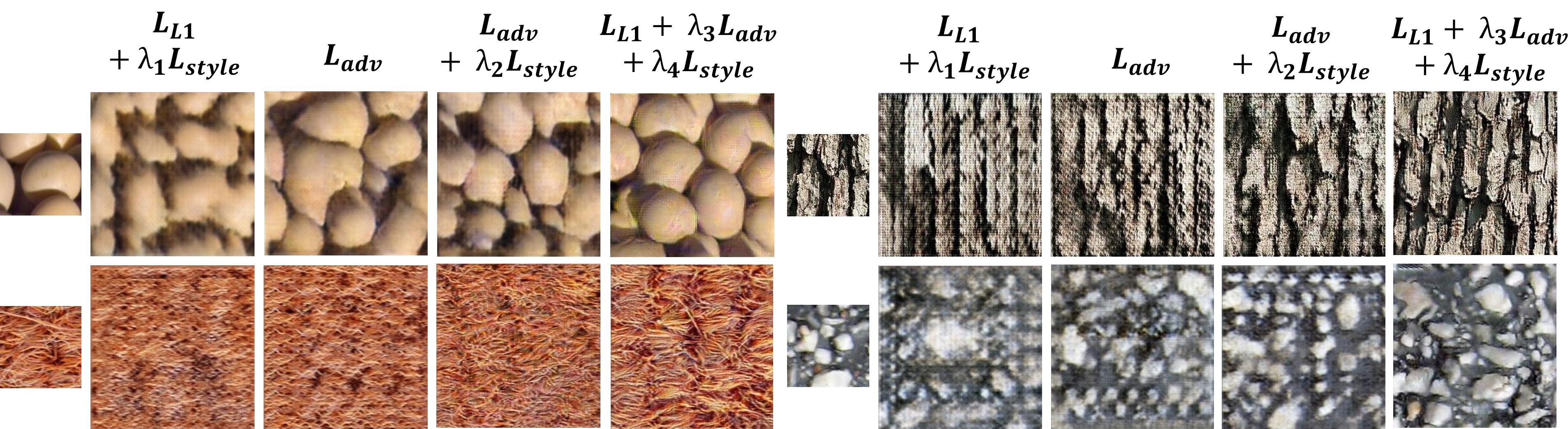}
\end{center}
   \caption{Images generated by the Texture Synthesis Network. The smaller patches denote the inputs to the Synthesis Network  and the remaining columns show results generated by using different combinations of loss functions.}
\label{tex_syn}
\end{figure*}

\begin{figure*}[t]
\begin{center}
\includegraphics[width=1.02\linewidth]{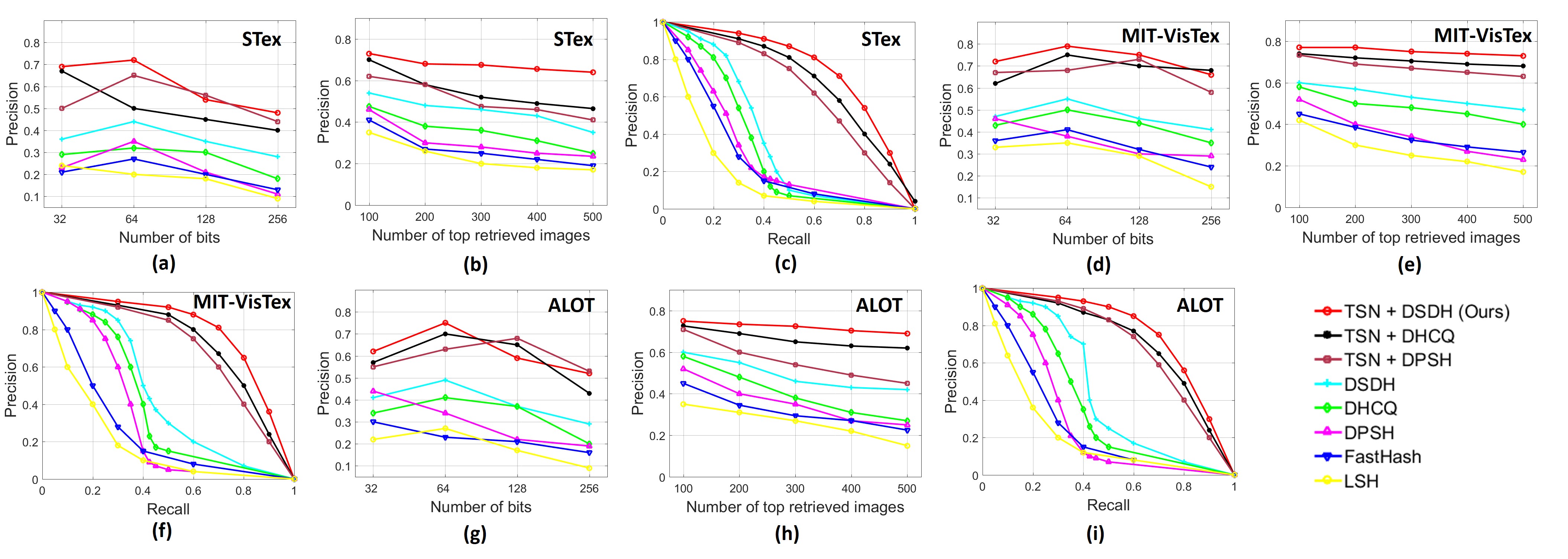}
\end{center}
   \caption{(a)-(c), (d)-(f), (g)-(i) show precision (Hamming radius $\leq$ 2) vs.\ number of bits, precision vs.\ number of top retrieved images, and precision vs.\ recall curve for STex, MIT-VisTex, and ALOT datasets respectively. }
\label{graphs}
\end{figure*}

\subsection{Datasets}

To demonstrate the efficiency of our method, we evaluated the proposed texture synthesis guided hashing framework on three popular texture datasets by comparing it against several state-of-the-art hashing frameworks. 
We use the \textbf{MIT-VisTeX} (full) database which is a collection of 167 unique texture patterns of size  $512 \times 512$ with one image belonging to every class. 
Next, we also show our experimental results on the Salzburg Texture \textbf{(STex)} database which is a collection of 476 unique texture patterns recorded under real-world conditions. Every image of this database is of size $1024 \times 1024$ with one image belonging to every category of texture. 
We select this database to test the performance of our framework for higher number of classes. 
We also show our experimental evaluations on the Amsterdam Library of Textures \textbf{(ALOT)} database that contains coloured images of 250 unique texture patterns. For every unique texture pattern, 100 images were recorded by varying various scientific parameters. 
Apart from this, linear mixtures of 12 materials are also included in the database totalling upto more than 27500 images in the database.
This database was chosen because it contains a larger intra-class variance which necessitates a robust framework for handling the same. 

\subsection{Implementation details} \label{4.2}
\begin{table*}[t]
\centering
\resizebox{17.5cm}{!}{
\begin{tabular}{|c||c|c|c|c||c|c|c|c||c|c|c|c|}
 \hline
 \multirow{2}{*}{Methods} & \multicolumn{4}{|c||}{STex} & \multicolumn{4}{|c||}{VisTex} &  \multicolumn{4}{|c|}{ALOT}\\
 \cline{2-13}
  & 32 bits & 64 bits & 128 bits & 256 bits & 32 bits & 64 bits & 128 bits & 256 bits & 32 bits & 64 bits & 128 bits & 256 bits\\
 \hline
 \hline
 TSN + DSDH (Ours) & \textbf{0.60} & \textbf{0.624} & \textbf{0.646} & \textbf{0.655} & 
 \textbf{0.731} & \textbf{0.746} & \textbf{0.759} & \textbf{0.762} & 
 \textbf{0.711} & \textbf{0.722} & \textbf{0.737} & \textbf{0.741}\\
 \hline
  TSN + DHCQ & 0.598 & 0.617 & 0.631 & 0.643 & 
  0.682 & 0.702 & 0.719 & 0.734 & 
  0.688 & 0.7 & 0.712 & 0.721\\
 \hline
  TSN + DPSH & 0.587 & 0.611 & 0.629 & 0.632 & 
  0.667 & 0.699 & 0.711 & 0.717 & 
  0.64 & 0.661 & 0.675 & 0.688\\
 \hline
  DSDH \cite{li2017deep} & 0.44 & 0.464 & 0.475 & 0.482 & 
  0.554 & 0.582 & 0.6 & 0.606 & 
  0.518 & 0.54 & 0.555 & 0.56\\
 \hline
  DHCQ \cite{tang2018supervised} & 0.432 & 0.46 & 0.471 & 0.478 & 
  0.542 & 0.559 & 0.573 & 0.581 & 
  0.451 & 0.49 & 0.527 & 0.542\\
 \hline
  DPSH \cite{li2015feature} & 0.430 & 0.420 & 0.415 & 0.413 & 
  0.498 & 0.512 & 0.52 & 0.524 & 
  0.43 & 0.458 & 0.483 & 0.498\\
 \hline
 FastHash \cite{lin2015supervised} & 0.369 & 0.381 & 0.394 & 0.396 & 
 0.387 & 0.414 & 0.431 & 0.449 & 
 0.324 & 0.354 & 0.386 & 0.41\\
 \hline
 LSH \cite{andoni2006near} & 0.347 & 0.363 & 0.377 & 0.382 & 
 0.368 & 0.395 & 0.411 & 0.419 & 
 0.303 & 0.351 & 0.37 & 0.376\\
 \hline
\end{tabular}}
\caption{MAP values of different methods using top-500 retrieved images.}
\end{table*}

\begin{table}[t]
\centering
\begin{tabular}{|c|c|c|c|c|}
 \hline
  \multicolumn{2}{|c|}{Variants} & STex & VisTex & ALOT\\
 \hline
 \multirow{4}{*}{\rotatebox[origin=c]{90}{TSN Loss}} & $L_{L1} + L_{style}$ & 0.19 & 0.23 & 0.213\\
 \cline{2-5}
 & $L_{adv}$ & 0.482 & 0.566 & 0.551\\
 \cline{2-5}
 & $L_{adv} + L_{style}$ & 0.586 & 0.671 & 0.643\\
 \cline{2-5}
 & $L_{adv} + L_{style} + L_{L1}$ & 0.624 & 0.696 & 0.722\\
 \hline
 \multicolumn{2}{|c|}{No CA} & 0.587 & 0.651& 0.678\\
 \hline
 \multicolumn{2}{|c|}{No Data Augmentation} & 0.556 & 0.592 & 0.659\\
 \hline
\end{tabular}
\caption{MAP values for different variants of the proposed model using code length of 64 bits.}
\end{table}

\paragraph{Data Preparation:} Since both MIT-Vistex and STex contain only one image for every texture class, we resize all the images from these datasets to $1024\times 1024$ and divide every image into 4 regions. Now randomly sampled patches covering three-fourth of the image region are used for training and rest one-fourth is used for sampling test patches, so that every testing sample remains unseen to the trained model. 
On the other side, ALOT has 100 samples for every texture class and we randomly select 70 images for training, 20 for testing and rest 10 for validation. This split is maintained for every experiment and we denote this division as $\{ D_{train}, D_{test}, D_{val}\}$. For our experiments, $D_{val}$ comes only from the ALOT dataset and all the hyperparameters are tuned based on this and remain constant for all the experiments. 
In our framework, there are two stages of training: we first train the Texture Synthesis Network (TSN) in an adversarial setup alongwith a discriminator using the objective in Equation \eqref{l1_loss}. 
Thereafter, we use the pre-trained TSN as a multi-scale feature extractor and train the second stage of the network with the objective in Equation \eqref{J} that finally outputs a $k$ bit binary vector for a given input texture patch. 
Following this, we use two types of training samples for training these two stages.  $\emph{Stage - 1}$: To train the TSN, we use paired data $\{I_{inp, stage1}^{128}, I_{gt, stage1}^{256}\}$, where $I_{gt, stage1}^{256}$ is an image of size $256 \times 256$ randomly sampled during every iteration of $\emph{Stage - 1}$ from $D_{train}$ and $I_{inp, stage1}^{128}$ is a randomly cropped $128 \times 128$ texture patch contained inside the $I_{gt, stage1}^{256}$. 
$\emph{Stage - 2}$: As the second stage is responsible for finally generating the binary hash codes, the second stage is supervised with the paired data $\{I_{inp, stage2}^{i}, L_{gt, stage2}^{i} \}_{i = 1}^{N}$, where  $I_{inp, stage2}^{i}$ is an image of size $128 \times 128$ cropped randomly from $D_{train}$ and  $L_{gt, stage2}^{i}$ denotes the corresponding class label of $i^{th}$ training sample. For every experiment, we generate a total of 2000 texture patches (from $D_{train}$) with respect to each class in order to train the network. During inference, we randomly generate 200 texture images (from $D_{test}$) of size $128 \times 128$ from every class  which are used as query to evaluate the performance. 

\paragraph{Training:} $\emph{Stage - 1}$: During the first stage of training, the generator $G_{TSN}(\cdot)$ takes $I_{inp, stage1}^{128}$ as input and outputs $I_{G}^{256}$ of size $256 \times 256$. The discriminator, $D_{TSN}(\cdot)$, tries to classify between the generated image $I_{G}^{256}$ (fake) and the corresponding ground truth $I_{gt, stage1}^{256}$ (real), and gives rise to adversarial loss, $L_{adv}$.  Reconstruction loss ($L_{L_{1}}$) is calculated by taking the absolute difference between $I_{G}^{256}$ and $I_{gt, stage1}^{256}$. We use a VGG-19 model pretrained on Imagenet to calculate the style loss ($L_{style}$) using Gram matrices computed at the output of the following layers : $relu1\_1, relu2\_1, relu3\_1, relu4\_1$ and $relu5\_1$.  
Following \cite{gatys2015texture}, a weighted summation of the corresponding losses is done by setting the weights to $0.244, 0.061, 0.15, 0.004, 0.004$ respectively.
A Gaussian distribution of mean 0 and standard deviation 0.02 is used to initialize the weights and biases of all the the convolutional layers. Following \cite{zhou2018non}, we use the Adam optimizer with initial learning rate 0.0002 and set momentum to 0.5.  We train the network for 100000 epochs, and $\gamma_{1}$ and $\gamma_{2}$ are set to 100 and 1 respectively. 
Once the training of $\emph{Stage - 1}$ is complete, we freeze all the weights of $G_{TSN}(\cdot)$ and use it as a feature extractor. $\emph{Stage - 2}$: In the second stage, we feed $I_{inp, stage2}^{i}$ to the TSN (i.e. $G_{TSN}(\cdot)$) and extract multi-scale feature representations from its intermediate layers. 
To alleviate the problem of insufficient data, we also use the generated texture patches $I_{G}^{256}$  from the output of TSN by sampling random patches of size $128 \times 128$, in order to train the second stage of network in an end-to-end manner.  
This aids in data augmentation and helps the second stage of the network to generalize well even from limited amount of original data. While training the second stage of the network, we follow the procedure used by the authors of \cite{li2017deep}.

\subsection{Performance Analysis}
 
In recent years, there has been a number of works addressing different loss functions to train end-to-end networks for hashing. DSDH \cite{li2017deep} employs a deep CNN-F architecture and uses pairwise label and classification information for generating hash codes. Similar to DSDH, DPSH \cite{li2015feature} uses pairwise label information, but learns feature representations and hash codes simultaneously. 
Similar to DPSH, DHCQ \cite{tang2018supervised} goes for simultaneous learning of feature representations and hash codes based on classification and quantization errors. 
Due to superior performance on various benchmark datasets, we have used the loss functions and training methods proposed in DSDH \cite{li2017deep} to train the second stage of our framework which finally outputs a $k$-bit binary vector. 
However, our generative model guided deep hashing framework is a meta-framework and can be easily extended with any other state-of-the-art's objective functions for training. 
Therefore, we have used objective functions from two other end-to-end trainable deep hashing frameworks \cite{li2015feature, tang2018supervised} in order to justify the efficacy of our framework. 
We name the experimental setups as TSN-DSDH, TSN-DPSH, TSN-DHCQ; where we train the second phase of our network, guided by a generative model, using the loss functions introduced in DSDH, DPSH and DHCQ respectively. 
We also compare with the original frameworks used in DSDH, DPSH, DHCQ which are broadly based on feed-forward convolutional neural network. 
We also compare against traditional hashing methods like FastHash \cite{lin2015supervised} and LSH \cite{andoni2006near} where every texture patch is represented by a standard LBP feature descriptor. 
Following other works \cite{li2017deep, tang2018supervised}, we evaluate Mean Average Precision (MAP) for different methods using same standard splitting protocol on three different datasets (see section \ref{4.2}). 
Table 1 depicts the MAP values at different lengths of hash code. In Figure \ref{graphs}, we have shown the plot for precision vs.\ top-T retrieved samples, precision vs.\ number of bits in hash code and precision vs.\ recall keeping the threshold for Hamming distance equal to 2. From Figure \ref{graphs}, it is evident that the performance of DSDH, DPSH, DHCQ is limited. 
Whereas, the same loss objective significantly boosts up the performance when combined with our TSN guided framework. 
Comprehensive experiments also validate that our method is the most time efficient with the average retrieval time per query (in seconds) being 0.00104, 0.00169 and 0.00281 for ALOT, MIT-VisTeX and STeX respectively.
From our observation, this significant improvement is mainly due to two reasons: first, our generative model guided framework can facilitate better feature learning, and second, the generated outputs of TSN help in data augmentation for the training of second stage of our framework.   
  
\subsection{Ablation Study}
 
In our framework, the central idea is to use a powerful generative model (TSN) to help generate better feature representations for texture image retrieval. 
Therefore, we carry out a comprehensive study to judge how image retrieval performance is related to the quality of image generation network. 
Figure \ref{tex_syn} depicts the quality of generated texture patches due to the use of different combinations of loss functions. In four different experiments, we train the TSN using the following combinations of the loss functions : a) $L_1$ loss and style loss (b) adversarial loss (c) adversarial and style loss (d) adversarial loss, style loss and $L_1$ loss. 
Thereafter, we use such pre-trained TSN networks as feature extractors in the second stage of training one at a time. 
We notice the performance due to pre-trained TSN using combination (a) to be very poor. 
The main driving force of this TSN network is the adversarial loss function. 
However, adding both L1 Loss and style loss alongside helps the TSN to generate better quality images, and thereby signifying a significant rise in the texture retrieval performance. 
The performance due to different possible combinations of loss objectives to train the TSN is shown in Table 2.
This implies that the better the generated images are, the more powerful are the feature representations preserved in the intermediate layers, thereby significantly improving the retrieval performance. 
Apart from considering the importance of individual losses of TSN, we also conduct experiments to verify the importance of channel-wise attention. 
We notice a drop of 0.037, 0.045 and 0.044 in the MAP values (with code length 64) for VisTex, STex and ALOT dataset respectively by removing the channel-wise attention in the second stage. 
We also conduct experiments without using data augmentation from the generated texture patches. Consequently, we notice a drop in the performance which is significantly large for  Vistex and STex datasets due to them having less amount of data.  
 
\section{Conclusion}
 
We have introduced a novel deep hashing architecture for texture image retrieval. 
Our framework first pre-trains a TSN which learns to synthesize an expanded view of a given texture thus recording texture specific information in its intermediate layers.
The binarized hash codes are finally obtained by gathering feature maps from these intermediate layers and combining them in a selective and progressive manner. 
We also alleviate the problem of limited training data by using the generated texture patches from the TSN for training. 
Thus we conclude that, the idea of extracting more robust features guided by generative networks can further be extended to other image retrieval problems.

{\small
\bibliographystyle{ieee}
\bibliography{egbib}
}

\end{document}